\documentclass[11pt]{article}

\usepackage[preprint]{acl}

\usepackage{times}
\usepackage{latexsym}

\usepackage[T1]{fontenc}

\usepackage[utf8]{inputenc}

\usepackage{microtype}

\usepackage{inconsolata}

\usepackage{times}  
\usepackage{helvet}  
\usepackage{courier}  
\usepackage{graphicx} 
\usepackage{booktabs, multirow, dcolumn}
\usepackage{algorithm}
\usepackage{algorithmicx}
\usepackage{algpseudocode}
\usepackage{amsmath}
\usepackage{amssymb}
\usepackage{amsfonts}
\usepackage{newfloat}
\floatstyle{ruled}
\newfloat{listing}{tb}{lst}{}
\floatname{listing}{Listing}
\usepackage{newfloat}
\usepackage{listings}
\usepackage{tikz}
\usepackage{tcolorbox}

\title{Exploring Layer-wise Information Effectiveness for Post-Training Quantization in Small Language Models}

\author{
 \textbf{He Xiao\textsuperscript{1}},
 \textbf{Qingyao Yang\textsuperscript{1}},
 \textbf{Dirui Xie\textsuperscript{2}},
 \textbf{Wendong Xu\textsuperscript{1}},
 \textbf{Zunhai Su\textsuperscript{1,3}}, \\
 \textbf{Runming Yang\textsuperscript{1}}, 
 \textbf{Haobo Liu\textsuperscript{1}},
 \textbf{Wenyong Zhou\textsuperscript{1}},
 \textbf{Zhengwu Liu\textsuperscript{1}},
 \textbf{Ngai Wong\textsuperscript{1}\footnote[1]{}}
\\
 \textsuperscript{1} The University of Hong Kong, 
 \textsuperscript{2} Huazhong University of Science and Technology \\
 \textsuperscript{3} Shenzhen International Graduate School, Tsinghua University
\\
 \small{
   \footnote[1]{} Corresponding author: \href{mailto:email@domain}{nwong@eee.hku.hk}
 }
}

\begin{document}

\maketitle

\begin{abstract}
    Large language models with billions of parameters are often over-provisioned: many layers contribute little unique information yet dominate the memory and energy footprint during inference. We present LieQ (\underline{L}ayer-wise \underline{i}nformation \underline{e}ffectiveness \underline{Q}uantization), a hardware-native, metric-driven post-training quantization framework that addresses the critical challenge of maintaining accuracy in \emph{sub-8B} models, model parameters less than 8B, under extreme low-bit compression. LieQ keeps \emph{uniform bit-width within each layer} while mixing precision across layers, preserving standard multiplication kernels and avoiding irregular memory access, codebooks, or irregular formats at inference time.
    Our method uncovers a strong correlation between layer-wise \emph{functional saliency} and \emph{representational compactness}, revealing that layers with higher training-induced energy concentration are functionally irreplaceable. Leveraging this insight, we propose a purely \emph{geometry-driven} sensitivity proxy that enables automatic bit-width allocation under a target average-bit budget without expensive gradient updates or inference-based perplexity probing. At sub 2-bit, LieQ consistently reduces the large accuracy gap typically observed for naive 2-bit baselines on Qwen3 and LLaMA3.x families, while retaining standard-kernel efficiency. These properties make LieQ a practical path toward deploying small language models on resource-constrained edge devices. Code will available here: \url{https://github.com/HeXiao-55/LieQ-official.git}.
\end{abstract}
    
\section{Introduction}
\label{sec:introduction}
\begin{figure}[tbp]
  \centering
  \includegraphics[width=0.45\textwidth]{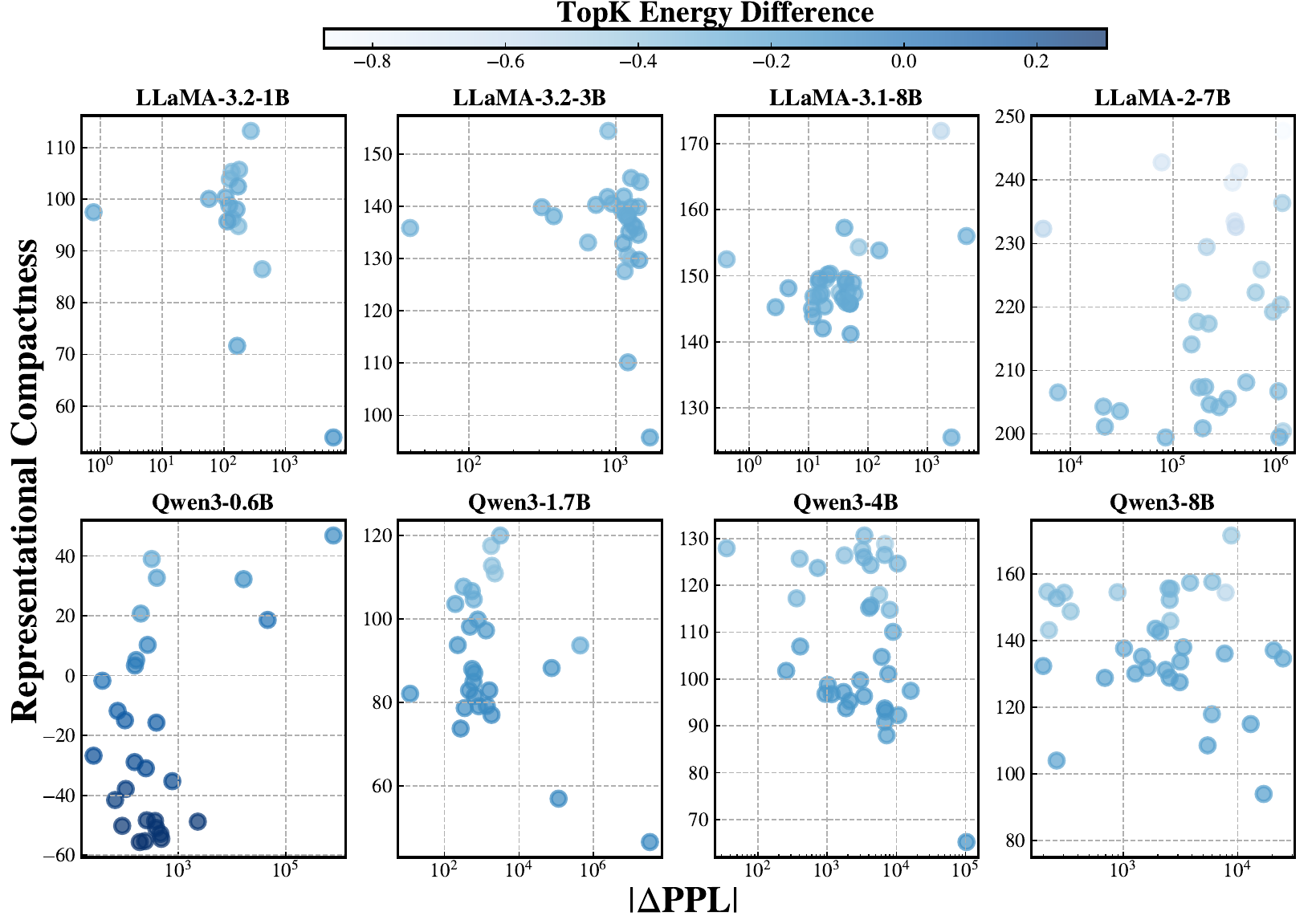}
  \caption{Layer-wise information taxonomy (each dot corresponds to one layer) across three correlated diagnostic metrics. Smaller models (e.g., Qwen-0.6B) exhibit lower robustness under extreme low-bit quantization, with certain layers being significantly more critical (clustered dots with deeper color) than others. Increasing the model size spreads out and balances the importance across layers.}
  \label{fig1}
\end{figure}
Large language models (LLMs) have achieved remarkable success across a variety of natural language processing tasks~\citep{achiam2023gpt,guo2025deepseekr1}, yet their immense parameter counts and activation sizes impose severe burdens on memory footprint and inference latency: a 7\,billion~(B) model fits into a single GPU, but 70\,B variants already demand 140\,GB of memory, far exceeding the less than 8\,GB budgets of widely deployed edge devices. Robotics applications also require lower-power small language models (SLMs) to run on robots/cars/drones with a constrained memory budget. 
SLMs are the future of generic edge AI~\citep{slm}. However, ~even models below 7\,B parameters, often marketed as ``mobile-friendly'', still surpass the 4 to 12\,GB memory envelopes of consumer phones and single-board computers, making aggressive compression an indispensable prerequisite for on-device inference~\citep{llama3,touvron2023llama2openfoundation,qwen2.5,qwen3}. This memory wall is not merely an engineering inconvenience; it is a fundamental barrier to democratizing LLMs. 

Quantization techniques offer a pathway to compress model weights and activations into low-bit representations, but in practice they struggle to maintain the original model accuracy. Quantization-aware training (QAT) can alleviate this but with the cost of substantial retraining overhead~\citep{Efficientqat}. 
Post-training quantization (PTQ) has therefore emerged as a promising remedy, compressing weights to low-bit without re-training~\citep{frantar-gptq,jlin2024AWQ}. However, PTQ often causes severe performance degradation at ultra-low bit-widths, with 2-bit quantization increasing perplexity by two orders of magnitude. \emph{This problem is especially severe in compact models (sub-8B parameters) due to their limited redundancy to absorb quantization noise.} Empirical studies confirm that the perplexity gap between full-precision and 2-bit representations narrows as the model size increases. In addition, existing low-bit approaches have limitations: sparse-outlier methods rely on heuristics for bit allocation~\citep{spqr}; salience-driven quantization maintains uniform bit budgets across layers~\citep{shang2023pbllm,huang2024billm}; and finer-grained methods like APTQ~\citep{aptq,silm} and LLM-MQ~\citep{mq} sacrifice hardware efficiency with heterogeneous data formats. This raises our first research question: \textit{\textbf{Challenge 1.) Can we employ a structured PTQ scheme that preserves accuracy while maintaining a regular weight layout?}}

Beyond precision loss, accuracy collapse also stems from the intrinsic model architecture. State-of-the-art (SOTA) LLMs predominantly follow the Transformer framework whose attention mechanism, while powerful, is notably fragile. Transformer layers exhibit heterogeneous saliency: some are highly vulnerable to precision loss, whereas others tolerate aggressive bit reduction. Prior work relies on costly ``drop-one-layer'' perplexity probing (functional saliency) to identify these critical layers. However, this functional behavior must stem from the underlying weight structures. This raises a fundamental question: \textit{\textbf{Challenge 2.) Can we identify a static geometric signature and theoretical guidance that explains and predicts functional saliency without pre-probing?}}

Finally, extreme low-bit quantization (2-bit) must confront memory-budget heterogeneity and hardware-efficiency tension. Edge accelerators, mobile SoCs and existing GPUs impose diverse memory ceilings, so a practical scheme must guaranty a target footprint while maximizing retained capability. Although per-vector or per-element mixed precision achieves high fidelity, it introduces irregular layouts, extra indices and costly decoding steps that offset theoretical gains~\citep{silm,huang2024billm,shang2023pbllm}. Codebook-based 2-bit methods also add runtime transforms that complicate deployment~\citep{aqlm,quarot,quip,omniq}. Hence, we ask: \textit{\textbf{Challenge 3.) How can we simultaneously attain hardware efficiency and accuracy robustness under extreme low-bit PTQ, while preserving standard kernels?}}

Therefore, targeting these challenges, we present LieQ (\underline{L}ayer-wise \underline{i}nformation \underline{e}ffectiveness \underline{Q}uantization), a principled PTQ framework. LieQ rests on a key analytical finding: \textbf{a layer's functional indispensability is strongly correlated with its representational compactness change.} 
We show that layers which undergo significant entropy reduction, learning to concentrate information into lower-rank manifolds, are the ones that cannot tolerate quantization noise. This insight allows us to use static geometric properties as a reliable proxy for saliency. Therefore, LieQ presents a main conclusion: \textbf{use high-precision(4-bit) for the layer with the most significant geometric feature (most compactness) and lower(2-bit) for others can retain performance to the maximum extent.} Our method dynamically allocates precision according to this geometric proxy, concentrating high-precision bits where they matter most while preserving hardware-friendly \emph{uniform} within-layer layouts and standard multiplication kernels. Our key contributions are:

1.\, We conduct a rigorous study connecting layer-wise functional saliency with representational geometry. We demonstrate that \emph{Compactness Shift}, the reduction in singular value entropy, serves as a strong predictor of a layer's true importance, offering an interpretability perspective on why certain layers break under quantization.

2.\,  We propose a purely geometry-driven quantization schedule. Instead of running expensive perplexity evaluations or tuning hyperparameter weights, LieQ selects high-precision layers solely based on their distribution properties. This yields a \emph{probing-free layer selection process} that is mathematically elegant and robust.

3.\,  We link this geometric ranking to a closed-form budget rule, achieving SOTA accuracy at $\sim$2.0 bits on Qwen3 and LLaMA3.x families while maintaining uniform-within-layer regularity for standard kernel compatibility.

\section{Related Works}

\paragraph*{Mixed-precision PTQ for Compact LLMs.} Early efforts such as GPTQ and RPTQ~\citep{frantar-gptq,rptq} reduced 175~B models to 4~bit with modest degradation, yet later studies revealed that the same routines collapse on sub-7B models where redundancy is scarce~\citep{infijanice}. Subsequent methods therefore inject protective mechanisms: outlier sparsity~\citep{spqr}, salience-driven grouping with mixed precision~\citep{silm}, and second-order error metrics with finer-grained mixed-precision~\citep{aptq,mq}. While these techniques restore language modeling accuracy at 2 to 4-bit, they typically rely on non-uniform formats that break tensor contiguity and hinder kernel fusion, motivating our search for a structured yet hardware-friendly scheme.

\paragraph*{Weights Effectiveness Diagnostics.} Transformer layers exhibit heterogeneous saliency: some are highly vulnerable to precision loss, whereas others tolerate aggressive bit reduction. To decide which layers deserve higher precision, prior work estimates sensitivity via weight reconstruction error~\citep{hawq}, Hessian eigenvalues~\citep{hawqv3}, activation entropy~\citep{ewq}, or direct layer-wise quantization~\citep{layerwisequantization}. Previous studies on systematic outliers \cite{an2025systematic,su2025kvsink} in LLMs have shown that certain shallow layers generate function-specific outliers and are particularly sensitive to model compression techniques \cite{yu2024super,su2025unveiling}. Recent probes measure representational geometry, i.e., rank expansion and spectral concentration~\citep{differank}, offering an alternative view to loss and accuracy.

\paragraph*{Bit-Width Allocation Algorithms.} Given per-layer scores, the remaining task is to satisfy a global memory budget based on hardware constraints. Greedy heuristics dominate early mixer-precision quantization papers, but their myopic choices often miss the ideals. Integer inference~\citep{knap} and differentiable search~\citep{dmpq} achieve better solutions at higher cost, whereas hardware-aware schedulers such as HAWQ-v3~\citep{hawqv3} trades optimality for throughput. Our approach inherits the efficiency of greedy search yet benefits from more informative scores, yielding a closed-form allocation for typical 2/3/4-bit settings.

\paragraph*{Extremely Low-Bit Quantization.} QuIP and QuIP\#~\citep{quip,quipsharp} introduce rotations (Hadamard) to improve incoherence and enable 2-bit quantization, while QTIP~\citep{qtip} leverages trellis coding and AQLM~\citep{aqlm} uses additive codebooks for extreme compression. These methods can deliver strong 2--3~bit accuracy but typically require additional process or runtime transforms and irregular memory access, complicating deployment. PTQTP~\citep{ptqtp} develops a more efficient ternary implementation of robust and plug-in PTQ method while performing impressive performance on mainstream datasets. They offer higher fidelity at moderate bit-widths, often trading hardware regularity and additional operations for accuracy. QuaRot~\citep{quarot} learns rotations to remove outliers for 4-bit, and SpQR~\citep{spqr} combines sparsity and quantization to approach nearly lossless compression.  In contrast, our approach targets the complementary regime of near-2-bit budgets while preserving standard multiplication kernels via uniform-within-layer layouts and mixes precision only across layers for ease of deployment. 



\begin{figure*}[t]
  \centering
  \includegraphics[width=0.9\textwidth]{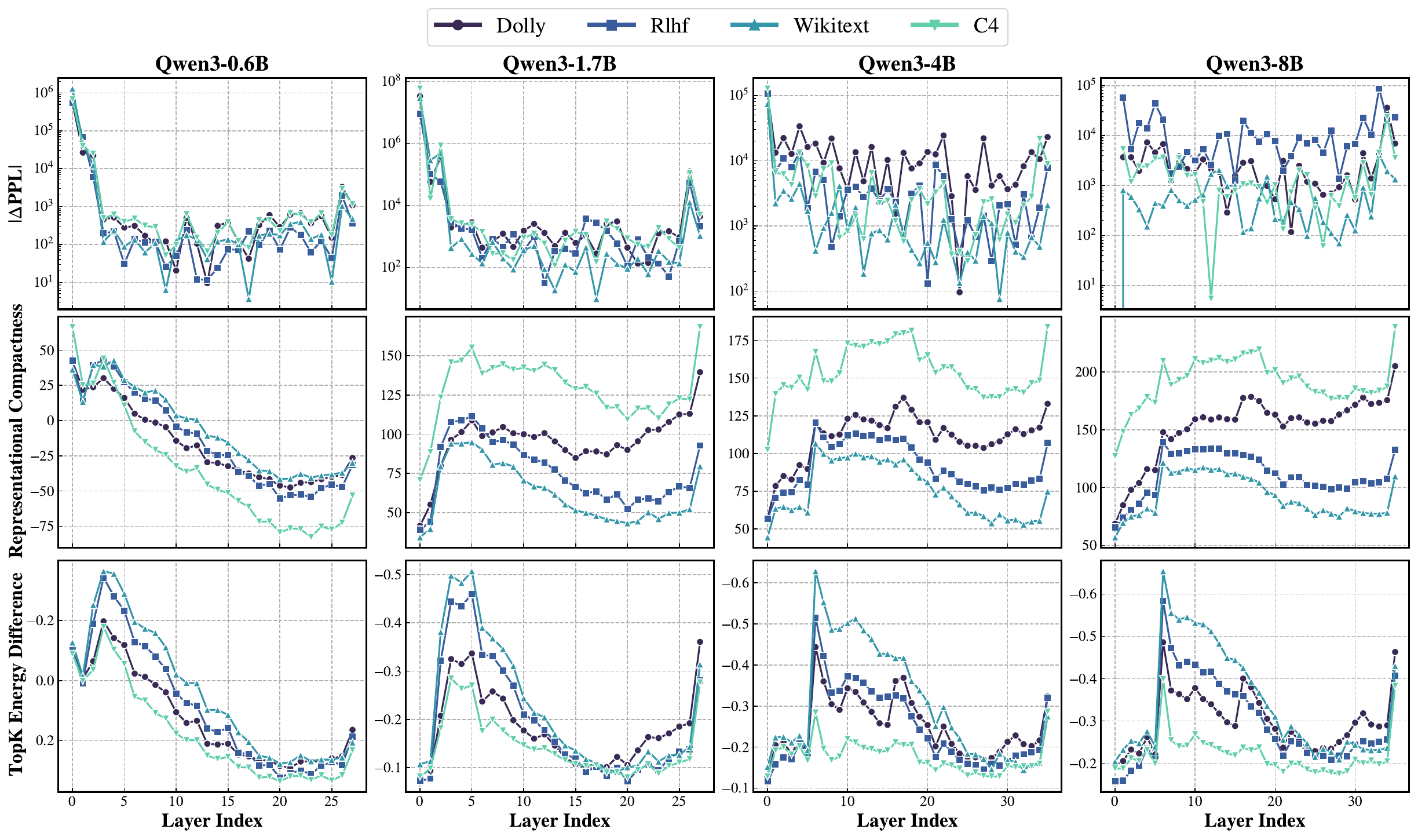}
  \caption{Functional diagnostic measures the drop in perplexity when a layer is removed on Qwen3 family. We use the representational compactness and TopK energy to proxy the significant perplexity loss.}
  \label{fig:qwen_metrics}
\end{figure*}

\section{Explore Stably Efficient PTQ Criterion}
The research landscape of efficient and robust PTQ can be mapped on solving the challenges highlighted in \textit{C1-C3} from three axes: (i) accuracy-compression trade-offs in \emph{sub-8B} PTQ, (ii) layer-wise diagnostics for guiding mixed precision, and (iii) automatically bit-width allocation under hardware constraints.

\subsection{Sensitivity Diagnostics}
Existing methods either secure accuracy at the expense of irregular layouts or retain regularity while ignoring layer heterogeneity. To address these challenges, we introduce a principled metric-driven quantization strategy that quantifies the saliency of each layer via complementary diagnostics. 

\textit{\textbf{Finding 1.} Smaller models (e.g., Qwen-0.6B) exhibit lower robustness under extremely low-bit quantization($\sim$2bit), with certain important layers standing out as significantly more critical than others. As illustrated in Figure~\ref{fig1}, increasing the model size leads to a more uniform distribution of representational compactness, resulting in a more balanced importance across layers. However, this also makes it harder to distinguish the relative significance of each layer.}

\paragraph*{Functional Saliency.} Our goal is to quantify the unique information contributed by each Transformer layer in an auto-regressive model \(\mathcal{M}\) composed of \(L\) total layers. Given a dataset \(\mathcal{D}=\{x^{(i)}\}_{i=1}^N\) containing \(N\) token sequences of length-\(T\), we define the baseline negative log-likelihood
\begin{equation}
  \mathcal{L}_{\mathrm{base}} = -\frac{1}{N}\sum_{i=1}^N\frac{1}{T}\sum_{t=1}^T \log p_{\mathcal{M}}(x_t^{(i)} \mid x_{<t}^{(i)})
\end{equation}
Where \(p_{\mathcal{M}}(x_t^{(i)} \mid x_{<t}^{(i)})\) is the probability of a token \(x_t\) given the preceding tokens \(x_{<t}\). The exponent of this loss yields the baseline perplexity, \(\mathrm{PPL}_{\mathrm{base}}=\exp(\mathcal{L}_{\mathrm{base}})\). By replacing the \(\ell\)-th Transformer block with an identity mapping plus its residual projection, we obtain the perturbed model \(\mathcal{M}_{\setminus\ell}\) and record its perplexity shift:
\begin{equation}
  \Delta\mathrm{PPL}_\ell = \mathrm{PPL}_{\setminus\ell} - \mathrm{PPL}_{\mathrm{base}}
\end{equation}
Where \(\mathrm{PPL}_{\setminus\ell}\) is the perplexity of the model without layer \(\ell\). This metric directly measures the drop in predictive performance attributable to the absence of layer \(\ell\). However, computing $\Delta\mathrm{PPL}$ requires $(L+1)N$ forward passes, making it computationally prohibitive for on-device or rapid deployment scenarios. This motivates the search for a geometric proxy.

\paragraph*{Representational Compactness.} We introduce a geometric proxy that quantifies layer saliency via Representational compactness properties. 
The theoretical foundation rests on the hypothesis that training enhances layer effectiveness by concentrating meaningful information into structured manifolds, which can be detected through changes in the singular value distribution.
For each type of linear projection \(P\) in the Transformer architecture, we analyze the projected representations \(Z = W_P^{(\ell)}\mathbf{h}^{(\ell)} \in \mathbb{R}^{T \times d_{\text{head}}}\). Here, \(W_P^{(\ell)} \in \mathbb{R}^{d_{\text{head}} \times d}\) represents the learned weight matrix for the respective projection type P in layer \(\ell\), transforming the input hidden states \(\mathbf{h}^{(\ell)} \in \mathbb{R}^{T \times d}\) into the corresponding space of attention head. To establish a baseline for comparison, we generate a randomly-initialized counterpart \(\tilde{Z} = \tilde{W}_P^{(\ell)}\mathbf{h}^{(\ell)}\), where \(\tilde{W}_P^{(\ell)}\) follows the same initialization distribution as the original weights but remains untrained.

The core insight is that training should increase the effectiveness of the layer by organizing information into tasks-relevant structures while eliminating noise and redundancy. To quantify this, we perform singular value decomposition on both matrices:
\begin{equation}
    Z = U\Sigma V^T; \quad \tilde{Z} = \tilde{U}\tilde{\Sigma}\tilde{V}^T
\end{equation}
yielding singular values \(\{\sigma_k\}\) and \(\{\tilde{\sigma}_k\}\), respectively. We then compute the representational compactness, defined as the exponential of the Shannon entropy of the energy distribution:
\begin{align}
\mathrm{Compact}(Z)& = \exp\Bigl(-\sum_{k=1}^{K} p_k\log p_k\Bigr) \\
p_k &= \frac{\sigma_k^2}{\sum_{j=1}^{K} \sigma_j^2}
\end{align}
Here, \(p_k\) represents the normalized energy of the \(k\)-th singular value in \(K=\min(T,d_{\text{head}})\) singular values. Representational compactness provides a smooth, differentiable measure of information concentration: when singular values are uniformly distributed, \(\mathrm{Compact}(Z)\) is high (indicating redundant representations), but when a few singular values dominate, \(\mathrm{Compact}(Z)\) is low (suggesting concentrated, sensitive representations). The layer effectiveness metric is then defined as the relative change in representational compactness:
\begin{equation}
\Delta r_\ell^{(P)} = \frac{\mathrm{Compact}(\tilde{Z}) - \mathrm{Compact}(Z)}{\mathrm{Compact}(\tilde{Z})}
\end{equation}
The normalization by \(\mathrm{Compact}(\tilde{Z})\) casts the metric as a relative change, making it a stable and comparable measure across different layers. It quantifies the proportional reduction in representational randomness, thus isolating the structural changes induced by training from baseline properties. A positive \(\Delta r_\ell^{(P)}\) indicates that training has increased layer effectiveness by developing more concentrated representations, making the layer more critical for model performance. 

\begin{table*}[t]
\centering
\small
\setlength{\tabcolsep}{4pt}
\begin{tabular}{@{}clcccccccc@{}}
\toprule
\multicolumn{2}{c}{\textbf{Qwen3}}  & \multicolumn{4}{c}{\textbf{Wiki}}                              & \multicolumn{4}{c}{\textbf{C4}}                          \\ 
\cmidrule(r){3-6} \cmidrule(r){7-10}
\begin{tabular}{l}\textbf{Weight} \\ \textbf{Precision}\end{tabular} & \textbf{Method} & \textbf{0.6B} & \textbf{1.7B} & \textbf{4B} & \textbf{8B} & \textbf{0.6B} & \textbf{1.7B} & \textbf{4B} & \textbf{8B} \\ 
\midrule
FP 16                 & -             & 20.9      & 16.7     & 13.64    & 9.71     & 30.31         & 22.36        & 19.83     & 15.41      \\
\midrule
\multirow{6}{*}{2bit} & GPTQ          & 2.38E+04  & 6.55E+02 & 5.92E+05 & 7.77E+04 & 1.70E+06   & 3.08E+06  & 4.37E+05  & 5.46E+05   \\
                      & AWQ           & 1.21E+07  & 7.52E+06 & 1.38E+07 & 1.21E+07 & 1.03E+09   & 7.70E+08  & 3.18E+08  & 1.37E+10   \\
                      & OmniQuant     & 2.55E+05  & N/A       & 5.30E+04 & 7.31E+04 & 1.78E+05   & N/A        & 5.81E+05  & 9.27E+04   \\
                      & PB-LLM        & 1.52E+05  & 9.27E+05 & 4.68E+03 & 1.80E+03 & 2.03E+05   & 2.21E+06  & 7.76E+03  & 2.28E+03   \\
                      & SliM-LLM      & 2.91E+04  & 4.41E+04 & 39.71    & N/A       & N/A         & N/A        & N/A        & N/A         \\
                      & \textbf{LieQ} & \textbf{36.19}     & \textbf{23.48}    & \textbf{18.48}    & \textbf{11.46}    & \textbf{43.44}      & \textbf{28.45}     & \textbf{26.72}     & \textbf{18.05}      \\ 
\midrule
\multirow{4}{*}{3bit} 
                      & AWQ           & 2.20E+02  & 84       & 91       & 27.5     & 45.93      & 27.15     & 32.93     & 17.38      \\
                      & OmniQuant     & 1.54E+05  & N/A       & 1.95E+05 & 4.59E+04 & 2.09E+05   & N/A   & 3.42E+05  & 4.73E+04   \\
                      & PB-LLM        & 2.47E+04  & 1.35E+05 & 3.23E+03 & 1.23E+03 & 3.03E+04   & 1.43E+05  & 2.28E+03  & 1.09E+02   \\
                      & \textbf{LieQ} & \textbf{26.02}     & \textbf{20.25}    & \textbf{18.19}    & \textbf{10.72}    & \textbf{36.89}      & \textbf{25.15}     & \textbf{25.61}     & \textbf{17.00}         \\ 
\bottomrule
\end{tabular}

\caption{Zero-shot perplexity (lower is better) on WikiText-2 and C4 across Qwen3 model sizes. \textbf{Bold}: best; \underline{underlined}: second-best. ``N/A'': not reported or not applicable.}
\label{table:ppl-qwen3}
\end{table*}

\paragraph*{Supporting Metric: Top-\(k\) Energy.} While compactness captures the overall distributional concentration, we also inspect the top-\(k\) energy fraction to confirm information concentration:
\begin{equation}
E_k(Z)=\frac{\sum_{i=1}^k \sigma_i^2}{\sum_j \sigma_j^2}, \quad \Delta E_{k,\ell}^{(P)} = E_k(Z) - E_k(\tilde{Z})
\end{equation}
A positive \(\Delta E_{k,\ell}^{(P)}\) signifies that training has shifted more energy into the dominant components. From our analysis, \(\Delta E\) closely tracks \(\Delta r\) and serves as a secondary validation of the geometric shift.
In our evaluation, we assess correlations on four diverse datasets: \text{WikiText2}~\citep{wiki}, \text{Dolly-15k}~\citep{dolly}, \text{HH-RLHF}~\citep{rlhf}, and \text{C4}~\citep{c4}. We observe strong Spearman correlation ($\rho > 0.8$) between \(\Delta\mathrm{PPL}_\ell\) and \(\Delta r_\ell\) across depths, confirming that geometric compactness is a robust proxy for functional sensitivity.

\textit{\textbf{Finding 2.} While different model families display distinct distributions of effective information, models within the same family consistently exhibit highly similar patterns across various test datasets, as shown in Figure~\ref{fig:qwen_metrics}. This intra-family consistency is observed across all three proposed metrics.}

Additionally, we observe that shallow layers tend to be more important, consistent with prior studies on super weights \cite{yu2024super,su2025unveiling}. 
Prior research has shown that certain shallow layers contain super weights, giving rise to systematic outliers \cite{an2025systematic} that reflect their mechanistic significance.
In both the Qwen and LLaMA series, our metrics consistently reveal the key functional importance of super weight layers, aligning with known mechanistic effects.

\begin{figure}[tbp]
  \centering
  \includegraphics[width=0.50\textwidth]{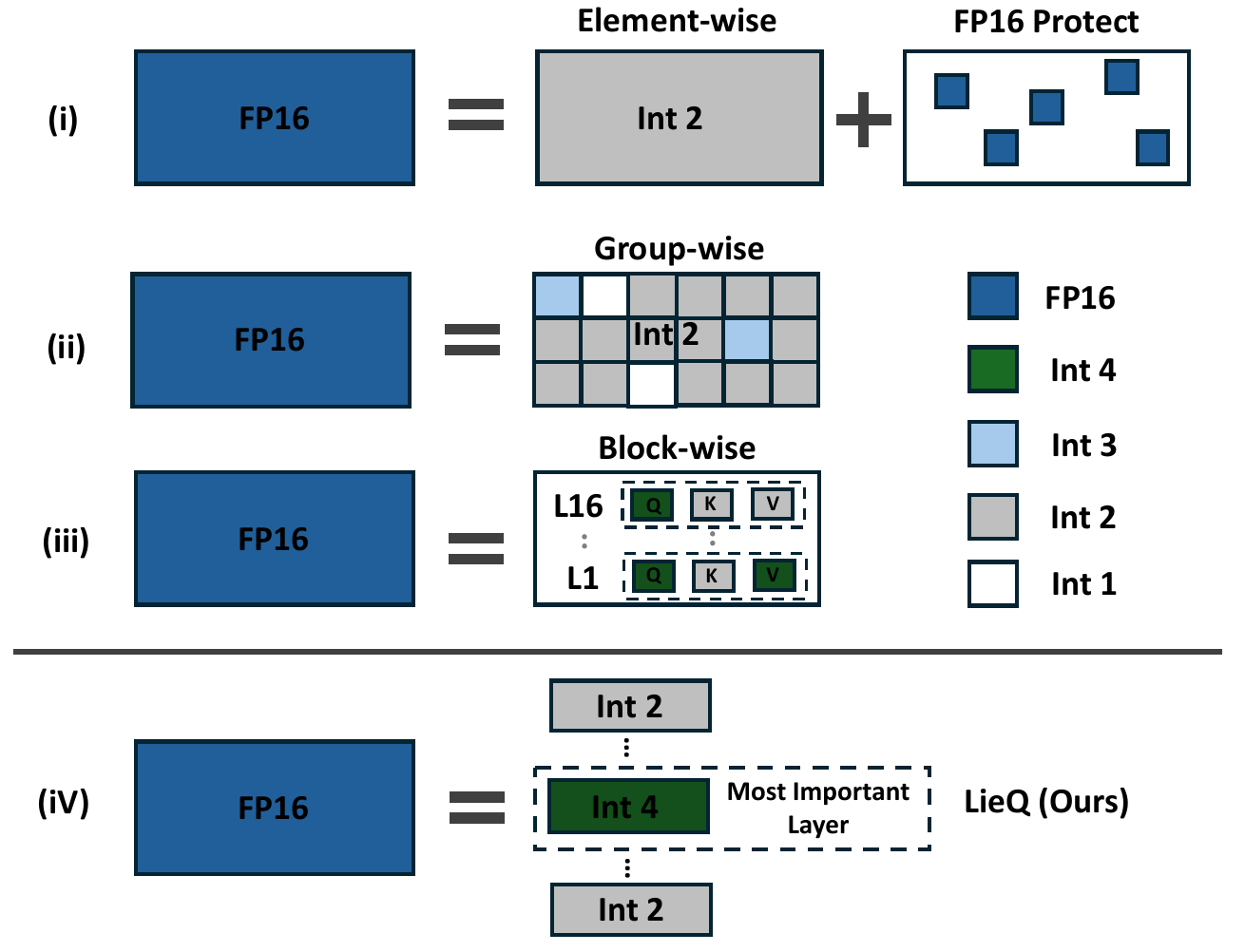}
  \caption{Illustration of the mixed-precision schemes. (i) Element-wise quantization with FP16 weights protection. (ii) Group-wise 2-bit quantization with 1-bit and 3-bit weights to balance accuracy and memory footprint. (iii) Block-wise 4-bit quantization within attention blocks in different layers. (iv) LieQ: Only one most significant layer with the most compact information is quantized to 4-bit, while the rest are quantized to 2-bit.}
  \label{fig:LieQ_Frame}
\end{figure}

\begin{table*}[htbp]
\centering
\small
\setlength{\tabcolsep}{5pt}
\begin{tabular}{@{}clcccccc@{}}
\toprule
\multicolumn{2}{c}{\textbf{LLaMA3}}                & \multicolumn{3}{c}{\textbf{Wiki}}          & \multicolumn{3}{c}{\textbf{C4}}              \\ 
\cmidrule(r){3-5} \cmidrule(r){6-8}
\begin{tabular}{l}\textbf{Weight} \\ \textbf{Precision}\end{tabular}      & \textbf{Method}            & \textbf{1B}        & \textbf{3B}        & \textbf{8B}        & \textbf{1B}         & \textbf{3B}        & \textbf{8B}         \\ 
\midrule
FP 16                 & -                 & 9.75      & 7.80      & 6.23      & 14.01      & 11.33     & 9.53       \\ 
\midrule
\multirow{7}{*}{2bit} & GPTQ              & 4.97E+03  & 2.42E+03  & 7.17E+02  & 1.96E+04   & \underline{2.69E+02}  & \underline{12.61}      \\
                      & AWQ               & 1.64E+05  & 1.11E+05  & 7.98E+05  & 1.78E+07   & 8.47E+05  & 1.89E+06   \\
                      & OmniQuant         & 7.71E+02  & \underline{5.38E+02}  & 2.08E+03  & 2.10E+03   & 1.36E+03  & 4.10E+03   \\
                      & PB-LLM            & 3.87E+03  & 1.01E+04  & 1.89E+03  & 1.86E+03   & 1.98E+03  & 1.17E+03   \\
                      & SliM-LLM          & \underline{3.31E+02}  & 8.19E+02  & \underline{31.52}     & \underline{6.92E+02}   & 2.90E+03  & 1.79E+02   \\
                      & \textbf{LieQ}     & \textbf{14.53}     & \textbf{9.78}      & \textbf{8.15}      & \textbf{20.70}      & \textbf{14.75}     & \textbf{13.18}      \\ 
\midrule
\multirow{5}{*}{3bit} & GPTQ              & 18.99     & 16.34     & 9.12      & \textbf{19.01}      & \underline{14.08}     & \textbf{10.18}      \\
                      & AWQ               & 34.81     & 13.22     & 10.80     & 23.77      & 15.23     & 12.57      \\
                      & OmniQuant         & \underline{15.79}     & \underline{9.91 }     & \underline{8.03}      & 25.49      & 15.98     & 13.59      \\
                      & PB-LLM            & 1.42E+03  & 4.97E+02  & 1.07E+03  & 8.84E+02   & 3.48E+02  & 6.45E+02   \\  
                      & \textbf{LieQ}     & \textbf{13.58}     & \textbf{9.21}      & \textbf{7.24}      & \underline{19.69}      & \textbf{13.86}     & \underline{11.41}      \\  
\bottomrule
\end{tabular}
\caption{Zero-shot perplexity (lower is better) on Wiki(Wikitext2) and C4 datasets across LLaMA3 models of different sizes and various quantization methods. \textbf{Bold}: best result; \underline{underlined}: second-best.}
\label{table:ppl-llama3}
\end{table*}

\subsection{Structured and Hardware-Friendly Mixed-Precision Quantization}

Given the strong correlation between functional saliency and representational geometry, we dispense with perplexity probing and complex weighted metrics. Instead, we propose \textbf{LieQ}, a purely geometry-driven quantization framework.

\paragraph*{Geometry-Driven Layer Selection.}
We aggregate the compactness diagnostic over linear projections to obtain a single score per layer:
\begin{equation}
s_\ell = \Delta r_\ell = \mathbb{E}_{P}\;\mathbb{E}_{\text{head}} \big[\Delta r_\ell^{(P,\text{head})}\big]
\end{equation}
This score \(s_\ell\) represents the ``geometric irreplaceability'' of layer \(\ell\). We rank layers by descending \(s_\ell\) to select the set of high-precision layers \(\mathcal{S}_{\text{hi}}\) and low-precision layers \(\mathcal{S}_{\text{lo}}\):
\begin{equation}
 \mathcal{S}_{\text{hi}} = \operatorname*{TopK}_{m}(s_1,\dots,s_L), \quad \mathcal{S}_{\text{lo}} = [L]\setminus\mathcal{S}_{\text{hi}}
 \label{eq:set}
\end{equation}
When \(\ell\in\mathcal{S}_{\text{hi}}\), we assign \(b_\ell= 4\); when \(\ell\in\mathcal{S}_{\text{lo}}\), \(b_\ell= 2\). This selection is \textbf{data-free} in the sense that it requires no validation set inference, relying only on the intrinsic weight statistics, i.e., via SVD on a single representative forward pass.

\paragraph*{Automatically Slipping Precision.} Given a target average-bit budget \(\bar{b}\in[2,4]\), the fraction of 4-bit layers under equal-sized layers is \(f=\frac{\bar{b}-2}{4-2}\), yielding a closed-form \(m=\operatorname{round}(fL)\). With unequal parameter counts \(N_\ell\), we select the smallest \(\mathcal{S}_{\text{hi}}\) (by descending \(s_\ell\)) such that \(\sum_{\ell\in\mathcal{S}_{\text{hi}}} N_\ell \ge f\sum_{\ell=1}^L N_\ell\).
Finally, we report the weights-only relative memory fraction w.r.t.\ FP16. Let $N_\ell$ denote the number of parameters in layer $\ell$, thus the compression ratio (CR) can be defined as below:
\begin{equation}
 \mathrm{CR}=\frac{\sum_{\ell=1}^{L} b_\ell N_\ell}{16\sum_{\ell=1}^{L} N_\ell},\label{eq:compression}
\end{equation}
so that memory reduction (weights-only) relative to FP16 is \(16/\bar{b}\) where \(\bar{b}=\frac{\sum_\ell b_\ell N_\ell}{\sum_\ell N_\ell}\).

\paragraph*{Integration with Existing Quantization Methods.}
LieQ is orthogonal to the choice of PTQ back-end, providing a plug-and-play guideline for exist PTQ methods. In our implementation, mixing the proposed diagnostic strategy with GPTQ-4bit or AWQ-4bit yields further compression while maintaining high accuracy. Although LieQ supports any budget-implied number of promoted layers, in order to probe the quantization limits we promoted the smallest subset (e.g., top-1 $s_\ell$ to 4-bit), quantizing the rest to 2-bit. This extreme configuration illuminates the minimum precision required to preserve model performance while maximizing compression, and it preserves standard multiplication kernels at inference time.

\begin{table*}[ht]
\centering
\small
\setlength{\tabcolsep}{5pt}
\begin{tabular}{l c c c c c c c c c}
\toprule
\textbf{Model} & \begin{tabular}{c}\textbf{Weight} \\ \textbf{Precision}\end{tabular} & \textbf{Method} & \textbf{PIQA} & \textbf{ARC-e} & \textbf{ARC-c} & \textbf{BoolQ} & \textbf{HellaSwag} & \textbf{Winogrande} & \textbf{MMLU} \\
\midrule
\multirow{9}{*}{Q3-4B} & FP16 & - & 74.97 & 80.47 & 50.51 & 85.14 & 52.24 & 65.82 & 68.25 \\
\midrule
 & 2.00 & GPTQ & 52.18 & 25.55 & 22.53 & 38.27 & 25.43 & 45.78 & 26.44 \\
 & 2.00 & AWQ & 53.75 & 25.76 & \underline{23.04} & 37.83 & 25.74 & \underline{50.43} & \underline{26.89} \\
 & 2.00 & OmniQuant & 52.77 & 26.3 & 22.1 & 37.82 & 25.93 & 50.11 & N/A \\
 & 2.00 & SliM-LLM & \underline{57.51} & \underline{39.39} & 19.62 & \underline{56.45} & \underline{31.29} & 49.25 & N/A \\
 & 2.05 & \textbf{LieQ} & \textbf{71.89} & \textbf{75.55} & \textbf{45.31} & \textbf{84.43} & \textbf{46.5} & \textbf{64.24} & \textbf{63.89} \\
 \midrule
  
 \multirow{4}{*}{Q3-4B}& 3.00 & GPTQ & 53.05 & 26.22 & 21.93 & 43.06 & 25.7 & 50.68 & 25.05 \\
& 3.00 & AWQ & \textbf{72.03} & \underline{69.15} & \underline{39.93} & \underline{79.66} & \textbf{47.11} & \underline{60.38} & \underline{61.42} \\
& 3.00 & OmniQuant & 53.37 & 23.75 & 20.56 & 37.82 & 25.95 & 50.59 & N/A \\
& 3.00 & \textbf{LieQ}  & \underline{71.27} &\textbf{ 71.84} & \textbf{42.15} & \textbf{84.53} & \underline{46.85} & \textbf{64.88} & \textbf{65.92} \\

\midrule
\multirow{9}{*}{L2-7B} & FP16 & - & 78.07 & 76.39 & 43.52 & 77.1 & 57.15 & 67.25 & 41.85 \\
\midrule
 & 2.00 & GPTQ & 58.71 & 38.01 & 22.44 & \underline{50.98} & 29.93 & 53.43 & 23.26 \\
 & 2.00 & AWQ & 49.73 & 26.53 & 20.99 & 37.83 & 26.14 & 49.8 & \underline{24.51} \\
 & 2.02 & QuIP\# & 71.38 & 55.56 & 28.84 & N/A & 42.94 & 62.43 & N/A \\
 & 2.02 & AQLM & 74.76 & 63.68 & 32.76 & N/A & 49.55 & 65.67 & N/A \\
 & 2.29 & AQLM & \underline{74.92} & \underline{66.5} & \underline{34.9} & N/A & \underline{50.88} & 62.43 & N/A \\
 & 2.05 & \textbf{LieQ} & \textbf{77.48} &\textbf{ 68.1} & \textbf{38.14} & \textbf{69.45} & \textbf{53.75} &\textbf{ 65.98} & \textbf{32.57} \\
 \midrule
 \multirow{4}{*}{L2-7B}& 3.00 & GPTQ & 76.01 & \textbf{72.9} & \textbf{40.44} & \textbf{74.5} & 54.2 & \textbf{67.96} & \underline{33.44} \\
 & 3.00 & AWQ & 76.66 & 52.65 & 38.82 & 67.58 & \textbf{70.66} & 65.43 & 32.78 \\
 & 3.04 & AQLM & \underline{76.88} & 65.06 & 38.4 & N/A & 54.12 & 63.54 & N/A \\
 & 3.00 & \textbf{LieQ} & \textbf{77.31} & \underline{67.93} & \underline{39.25} & \underline{70.7} & \underline{55.01} & \underline{65.98} & \textbf{34.39} \\

\midrule

\multirow{8}{*}{L3-3B} & FP16 & - & 76.82 & 74.41 & 42.58 & 72.57 & 55.34 & 69.06 & 54.08 \\

\midrule
 & 2.00 & GPTQ & 53.37 & 26.94 & 20.65 & \underline{44.56} & \underline{26.81} & 50.28 & 23.66 \\
 & 2.00 & AWQ & 52.61 & 24.79 & \underline{22.27} & 37.83 & 25.38 & 49.64 & \underline{26.89} \\
 & 2.00 & OmniQuant & \underline{55.27} & \underline{29.54} & 18.6 & 37.83 & 26.3 & 50.27 & N/A \\
 & 2.00 & SliM-LLM & 53.48 & 26.43 & 19.71 & 38.62 & 26 & \underline{50.75} & N/A \\
 & 2.07 & \textbf{LieQ} & \textbf{75.41} & \textbf{70.33} & \textbf{37.71} & \textbf{64.5} & \textbf{50.88} &\textbf{ 68.11} & \textbf{48.25} \\
 \midrule
 \multirow{4}{*}{L3-3B} & 3.00 & GPTQ & 72.85 & \underline{67.63} & 36.6 & 65.5 & \underline{51.39} & 64.93 & 41.28 \\
 & 3.00 & AWQ & \underline{74.1} & 65.74 & \underline{36.69} & \underline{72.32} & 50.13 & \underline{66.33} & \underline{46.51} \\
 & 3.00 & OmniQuant & 73.61 & 65.06 & 34.38 & 67.33 & 49.39 & 63.3 & N/A \\
 & 3.00 & \textbf{LieQ} & \textbf{75.35} & \textbf{72.18} & \textbf{40.19} & \textbf{72.32} & \textbf{52.34} & \textbf{66.93} & \textbf{49.62} \\

\bottomrule

\end{tabular}
\caption{Comparison (higher is better) on seven zero-shot reasoning tasks. \textbf{Bold}: best; \underline{underlined}: second-best. ``N/A'': not reported or not available in the original source.}
\label{bench1}
\end{table*}

\section{Experiments and Results}

\paragraph*{Evaluation Protocol.}
We implemented our evaluation on PyTorch using models from HuggingFace~\citep{hf}. No task-specific calibration, or post-training fine-tuning was applied in any of our experiments. All of our experiments were conducted on a single NVIDIA RTX 3090 (24\,GB) GPU with mixed precision enabled and gradient checkpointing disabled to preserve activation needed for rank analysis. We follow AWQ~\cite{jlin2024AWQ} acceleration setting to establish the evaluation of end-to-end performance, the sequence length are settled to 512 for peer comparison.

\paragraph*{Models and Baselines.}
We evaluated the layer-wise information effectiveness of multiple mainstream LLM families including Qwen3 (Q3)~\citep{qwen3}, LLaMA3.x (L3)~\citep{llama3}, and LLaMA1\&2 (L1\&L2) ~\citep{touvron2023llama2openfoundation}. All evaluated model sizes are less than 7--8\,B parameters. We compare against representative PTQ methods: SliM-LLM~\citep{silm}, AWQ~\citep{jlin2024AWQ}, GPTQ~\citep{frantar-gptq}, and OmniQuant~\citep{omniq}.  We additionally position against rotation/codebook-based approaches reported in the literature, including QuIP/QuIP\#~\citep{quip,quipsharp} and AQLM~\citep{aqlm}, noting their differing deployment characteristics.

\paragraph*{Evaluation Datasets and Metrics.}
For language modeling quality, we report perplexity on WikiText-2~\citep{wiki} and C4~\citep{c4}. To assess cross-domain generalization and language reasoning performance, we evaluate LieQ and existing methods on ARC-C/E~\citep{arc-c}, BoolQ~\citep{boolq}, HellaSwag~\citep{hellaswag}, PIQA~\citep{piqa}, Winogrande~\citep{winog}, and MMLU~\citep{mmlu}, following established protocols~\citep{eval-harness}.

\begin{figure}[tb]
  \centering
  \includegraphics[width=0.4\textwidth]{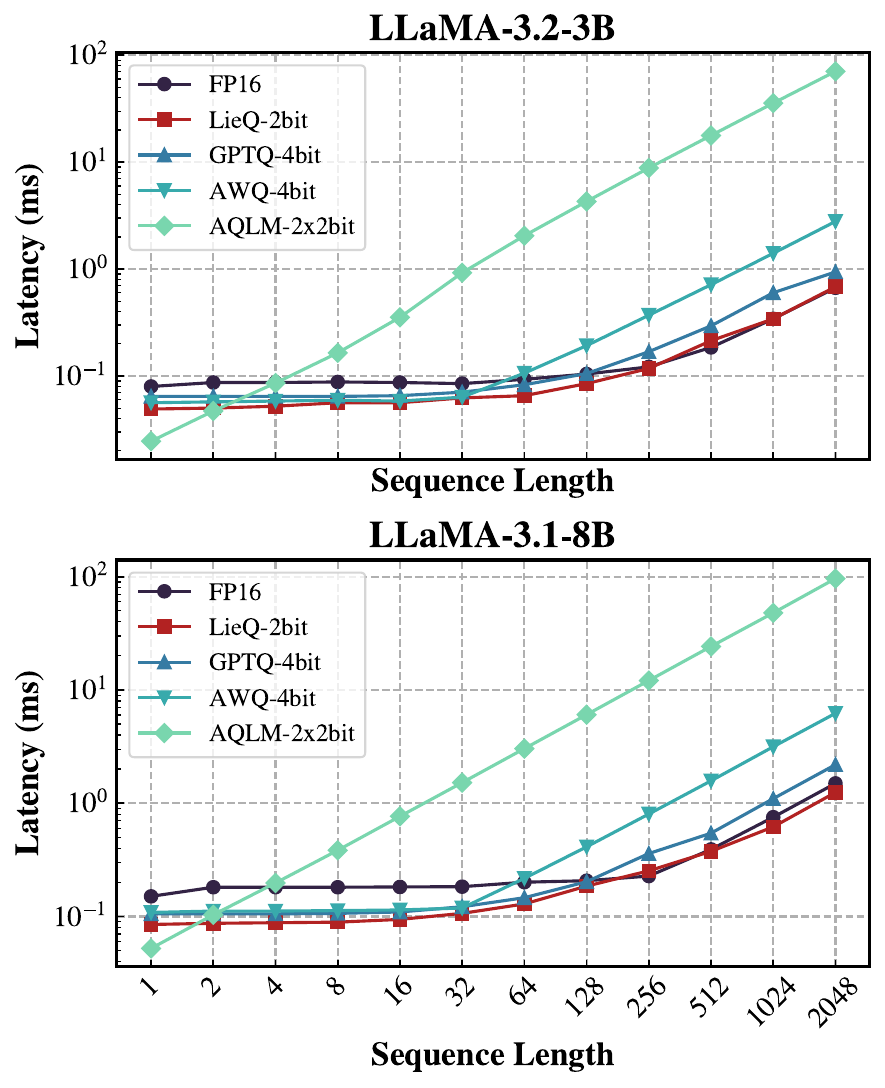}
\caption{Microbenchmark latency of the gate\_proj layer for LLaMA-3.2-3B and LLaMA-3.1-8B.}
  \label{fig:latency_gate}
\end{figure}

\paragraph*{Results and Analysis.}
The primary aim of LieQ is to provide solid analysis and clear guidelines for structured extreme low-bit PTQ. In addition, using our proposed diagnostic methods, we have observed two key performance advantages of LieQ:
~\textbf{i) Competitive accuracy in near-2-bit budgets.} As illustrated in Tables~\ref{table:ppl-qwen3} and \ref{table:ppl-llama3} and summarized in Table~\ref{bench1}, LieQ substantially mitigates the degradation seen in naive 2-bit baselines on both perplexity and zero-shot tasks for Qwen3 and LLaMA 3.x, and achieves SOTA performance. While several 3-bit methods could reach higher peak accuracy in some settings, LieQ attained competitive results without introducing runtime rotations or codebooks.
~\textbf{ii) Hardware-friendly regularity.} With uniform bit-width \emph{within} each layer, tensors remain contiguous and a single multiplication kernel per layer suffices. Figure~\ref{fig:latency_gate} presents a per-layer microbenchmark on the gate projection of LLaMA-3.2-3B and LLaMA-3.1-8B, indicating latency reduction relative to FP16 under identical experimental conditions. Our focus here on acceleration is to preserve standard kernels and avoid irregular formats that fragment GPU throughput. Moreover, we implemented and evaluated the method end-to-end on the Alpaca dataset using open-source implementations; the results are presented in Table~\ref{tab:e2e}. 

\begin{table}[t]
    \centering
    \small
    \setlength{\tabcolsep}{4pt}
    \begin{tabular}{l|c|c|c|c}
    \toprule
 Models  & \begin{tabular}{cc} Baseline \\ (Token/s) \end{tabular}  & \begin{tabular}{cc} Baseline \\ (GB) \end{tabular}  & \begin{tabular}{cc} \textbf{LieQ} \\ (Token/s) \end{tabular} & \begin{tabular}{cc} \textbf{LieQ} \\ (GB) \end{tabular} \\
    \midrule
    L1-7B  & 38.85 & 12.68 & 50.89 & 4.32 \\
    L1-13B & OOM   & 26.03 & 40.09 & 7.83 \\
    L2-7B  & 39.73 & 12.86 & 50.67 & 4.32 \\
    \bottomrule
    \end{tabular}
    \caption{End-to-end decode speed evaluation and comparison}
    \label{tab:e2e}
\end{table}

\begin{figure}[tbp]
  \centering
  \includegraphics[width=0.45\textwidth]{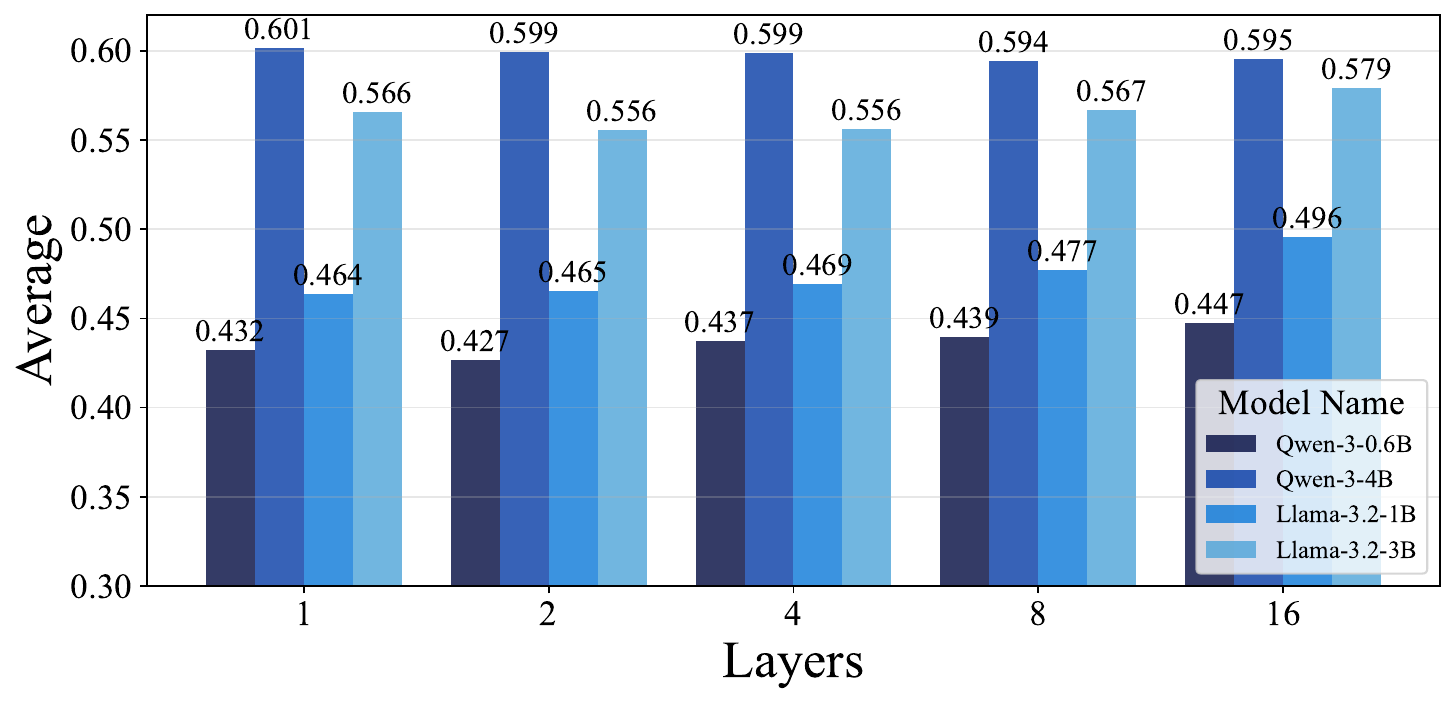}
  \caption{Average accuracy difference on language reasoning tasks with various precision configurations on small language models.}
  \label{ablation}
\end{figure}

\paragraph*{Sensitivity to Bit Budget.} Our experimental results demonstrate that LieQ maintains both excellent accuracy and inference performance on SLMs when using the structured, hardware-friendly mixed-precision framework. Furthermore, to explore the trade-off between the number of high-precision layers and model performance, we adjusted the budget configurations to conduct the ablation study on the automatic bit allocation method. Figure~\ref{ablation} shows the average language reasoning performance as we increased the number of 4-bit quantized layers (selected by our geometric proxy) from 1 to 16; this range covers precisions from 2-bit to 3-bit levels. This confirms that protecting even the smallest subset (one layer) of geometrically critical layers yields a significant recovery in accuracy.

\section{Conclusion}
In this work, we first identified a strong correlation between a layer's functional saliency and its representational geometry, revealing that layers with high energy concentration are structurally irreplaceable.
Building on this discovery, we proposed LieQ, a hardware-native PTQ framework that utilizes this geometric property as a proxy for bit-width allocation.
Our experiments on Qwen3 and LLaMA3.x confirm that LieQ effectively identifies critical layers, substantially mitigating the accuracy collapse of naive 2-bit baselines while preserving hardware efficiency.
These results not only offer a practical compression tool but also suggest that quantization sensitivity is a fundamental structural property rooted in weight manifolds, rather than a stochastic phenomenon.

\section*{Limitations}
Our goal is to establish a analysis proxy for guiding extremely low-bit quantization. Therefore, although we have evaluated inference efficiency via a simple end-to-end estimate, significant room remains for optimization in practical engineering. End-to-end inference throughput optimization will depend on specific system-level implementations. Nevertheless, we believe this exploration path is applicable to LieQ (thanks to our simple, symmetrical core design) and we will pursue further optimization in future engineering work.

\bibliography{custom}

\appendix
\begin{figure*}[t]
  \centering
  \includegraphics[width=\textwidth]{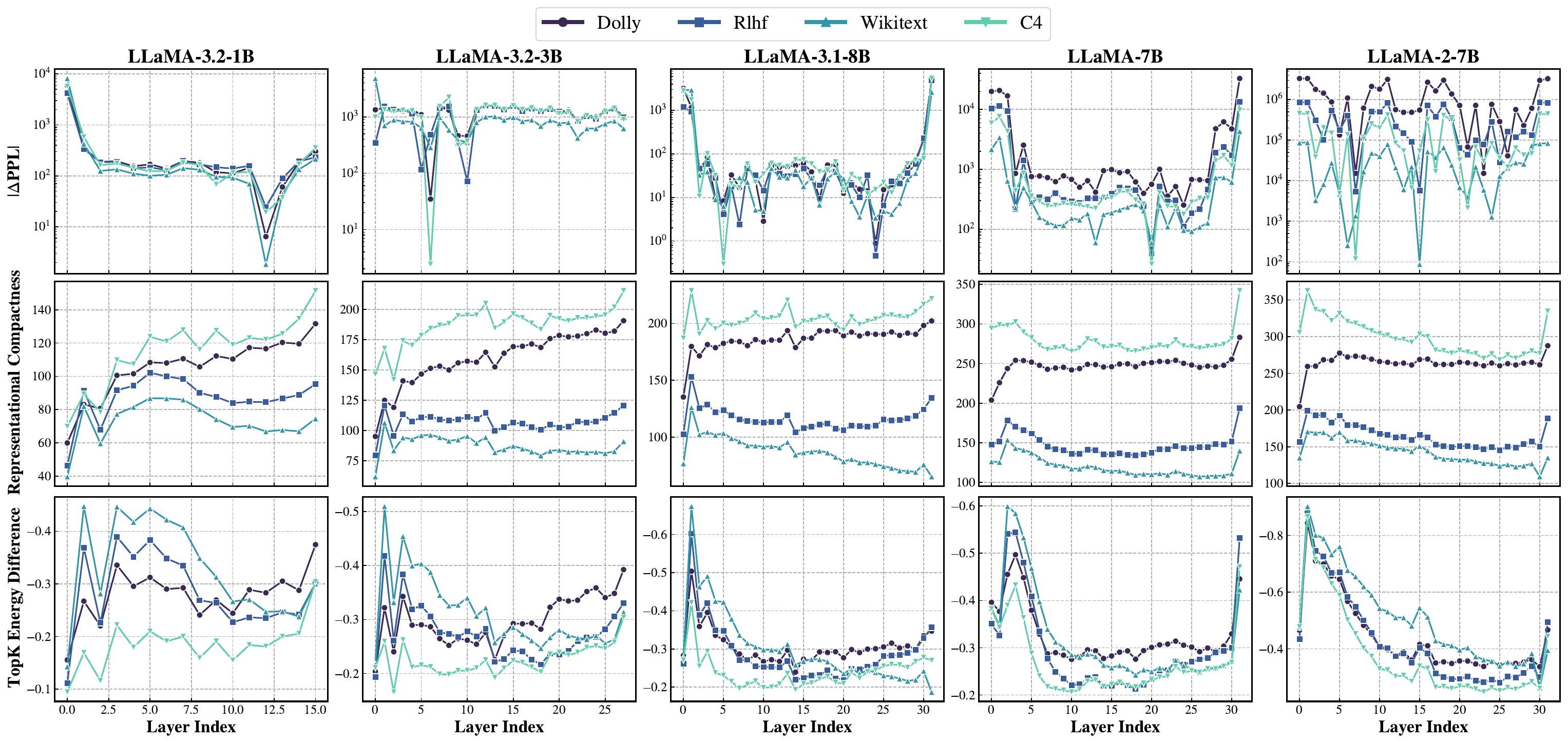}
  \caption{Functional diagnostic measures the drop in perplexity when a layer is removed on LLaMA3 family.}
  \label{fig:llama_metrics}
\end{figure*}
\section{More Result}
As illustrated in Figure~\ref{fig:llama_metrics}, we further shows the measurement for LLaMA3.X family models using LieQ and its diagnostic strategy.
\subsection*{LLM Usage Disclosure}
This paper is primarily the work of the human authors, but we also made use of several advanced LLMs, including ChatGPT-5 and Kimi-K2. They were employed to support result analysis, and help with formatting and language polishing. We acknowledge the contributions of these LLMs, while fully recognizing their limitations, and take full responsibility for all content presented under our names. We did not include hidden prompt-injection text in the submission, and all external data and code comply with their respective licenses.

\end{document}